\documentclass[runningheads]{llncs}
\usepackage[T1]{fontenc}
\usepackage{graphicx}
\begin{document}
\title{
Enhancing AI-Driven Psychological Consultation: Layered Prompts with Large Language Models
}
\titlerunning{Enhancing AI-Driven Psychological Consultation}
%
\author{Rafael Souza, Jia-Hao Lim, Alexander Davis
}
\authorrunning{Rafael Souza, Jia-Hao Lim, Alexander Davis
}
%
\institute{University of Brasilia
}
\maketitle              
\begin{abstract}
Psychological consultation is essential for improving mental health and well-being, yet challenges such as the shortage of qualified professionals and scalability issues limit its accessibility. To address these challenges, we explore the use of large language models (LLMs) like GPT-4 to augment psychological consultation services. Our approach introduces a novel layered prompting system that dynamically adapts to user input, enabling comprehensive and relevant information gathering. We also develop empathy-driven and scenario-based prompts to enhance the LLM's emotional intelligence and contextual understanding in therapeutic settings. We validated our approach through experiments using a newly collected dataset of psychological consultation dialogues, demonstrating significant improvements in response quality. The results highlight the potential of our prompt engineering techniques to enhance AI-driven psychological consultation, offering a scalable and accessible solution to meet the growing demand for mental health support.
\end{abstract}

\section{Introduction}

Psychological consultation is a crucial service aimed at improving mental health and well-being by providing individuals with professional guidance and support. The significance of psychological consultation lies in its ability to address a wide range of mental health issues, such as anxiety, depression, and stress, which are increasingly prevalent in today's society \cite{psyllm2023}. Effective psychological consultation can lead to improved emotional resilience, better coping strategies, and enhanced quality of life for individuals facing mental health challenges \cite{cpyscoun2024}.

Despite its importance, there are several challenges associated with psychological consultation. One of the primary challenges is the shortage of qualified mental health professionals, which limits the accessibility of these services, particularly in underserved regions \cite{psyllm2023}. Additionally, the traditional one-on-one consultation model is often not scalable, making it difficult to meet the growing demand for mental health support \cite{psyllm2023}. Furthermore, there is a need for maintaining privacy and confidentiality in online consultations, which adds another layer of complexity to delivering effective mental health services \cite{chatcounselor2023}.

Our motivation stems from these challenges and the potential of leveraging large language models (LLMs) to augment psychological consultation services. Recent advancements in LLMs, such as GPT-4, have shown promise in generating human-like responses and understanding complex human emotions \cite{gpt4-2023}. However, existing models often struggle with maintaining the emotional nuance required in therapeutic contexts and ensuring comprehensive understanding of diverse psychological issues. To address these limitations, we propose a novel approach that integrates refined prompt engineering techniques with LLMs to enhance their performance in psychological consultation tasks.

Our approach involves developing a series of dynamic prompts that adapt based on user input, utilizing a layered prompting system. This system starts with broad, open-ended questions to gather initial user concerns, followed by specific, context-sensitive prompts designed to delve deeper into the user's issues. For instance, initial prompts might include, "Can you tell me more about what's been on your mind lately?" followed by tailored prompts like, "How have these thoughts affected your daily life?" and "Have you noticed any patterns or triggers for these feelings?" This method ensures that the LLM gathers comprehensive and relevant information before offering advice.

Additionally, we implement empathy-driven prompts that encourage the LLM to respond with compassion and understanding. Prompts such as "That sounds really challenging, can you share more about how you're coping?" and "It's okay to feel this way, let's explore what might help you feel better" guide the LLM in maintaining a supportive tone. Furthermore, we incorporate scenario-based prompts, where the LLM is given specific case studies and example interactions during training, to help it better understand and simulate real-life counseling scenarios.

To evaluate the effectiveness of our approach, we conducted experiments using a newly collected dataset of psychological consultation dialogues. This dataset includes a diverse range of mental health issues and user interactions, ensuring that the model is trained on realistic and varied scenarios. We used GPT-4 for evaluation, assessing the model's ability to provide accurate, empathetic, and contextually appropriate responses. The results demonstrated significant improvements in the quality of the LLM's responses, highlighting the potential of our prompt engineering techniques in enhancing AI-driven psychological consultation \cite{whoischatgpt2024}.

\begin{itemize}
\item We propose a novel layered prompting system that dynamically adapts to user input, ensuring comprehensive and relevant information gathering for psychological consultation.
\item We develop empathy-driven and scenario-based prompts to enhance the emotional intelligence and contextual understanding of LLMs in therapeutic settings.
\item We validate our approach through rigorous experiments using a newly collected dataset and GPT-4 evaluations, demonstrating significant improvements in response quality.
\end{itemize}

\section{Related Work}

\subsection{Large Language Models}
The rapid development of deep learning has brought significant changes and advancements to the fields of computer vision \cite{zhou2023improving,wang2024memorymamba} and natural language processing \cite{zhou2022eventbert}.
Large Language Models (LLMs) \cite{zhou2024visual} have significantly advanced the field of natural language processing (NLP) due to their ability to understand and generate human-like text. Recent surveys and overviews highlight the architectural innovations, training strategies, and applications of LLMs such as GPT-3, GPT-4, PaLM, and LLaMA \cite{overview2023,survey2023}. These models have demonstrated remarkable capabilities across various tasks, including text generation, summarization, and translation, by leveraging vast amounts of training data and sophisticated neural architectures \cite{mplug2023,survey2-2023,zhou2022claret}.

The impact of LLMs extends to domains such as information retrieval, where they enhance the precision and expressiveness of user queries and improve retrieval efficiency \cite{ir-survey2023,zhou2021modeling,zhou,zhou2024fine}. Additionally, LLMs are being utilized in optimization tasks, serving as powerful tools for mathematical problem-solving and decision-making processes \cite{optimizers2023}. Their versatility is further evident in multimodal applications, where models integrate visual and textual data to perform complex tasks \cite{mplug2023,multimodal-survey2023}.

Despite these advancements, LLMs face challenges related to ethical considerations, such as bias, toxicity, and privacy concerns \cite{alignment-survey2023}. Researchers are actively exploring methods to align LLMs with ethical standards and mitigate these risks, ensuring their safe and responsible deployment \cite{alignment-survey2023,ethical2023}.

\subsection{Psychological Consultation}

The application of LLMs in psychological consultation is a growing area of research, driven by the need to address mental health issues efficiently and effectively. Traditional psychological counseling methods, which rely heavily on human interaction, face challenges such as scalability, accessibility, and privacy concerns \cite{psyllm2023,chatcounselor2023}. The integration of LLMs into psychological services aims to overcome these limitations by providing scalable and accessible mental health support.

Several studies have proposed frameworks and models that leverage LLMs for psychological consultation. For instance, CPsyCoun focuses on reconstructing and evaluating multi-turn dialogues to create a high-quality dataset for training LLMs in the context of psychological counseling \cite{cpsycoun2023}. Similarly, Psy-LLM and ChatCounselor utilize LLMs to provide online mental health support, enhancing the accessibility and affordability of psychological services \cite{psyllm2023,chatcounselor2023}.

Innovative approaches such as K-ESConv and BianQue inject professional knowledge into LLMs, improving the quality and diversity of responses in emotional support dialogues \cite{kesconv2023,bianque2023}. These methods highlight the potential of LLMs to offer empathetic and contextually appropriate support, addressing the emotional and psychological needs of users effectively.

Overall, the integration of LLMs into psychological consultation presents a promising solution to the growing demand for mental health support, offering scalable, accessible, and effective services. However, ongoing research is necessary to address ethical considerations and ensure the safe deployment of these technologies in sensitive contexts such as mental health.

\section{Dataset}

In this section, we describe the collection and processing of the dataset used in our experiments, followed by an explanation of the evaluation metrics employed, with a particular focus on the innovative use of GPT-4 as an evaluator, moving away from traditional metrics.

\subsection{Data Collection}

To ensure a diverse and comprehensive dataset, we collected psychological consultation dialogues from multiple online platforms and mental health forums. The data collection process involved the following steps:

\begin{itemize}
    \item \textbf{Source Selection}: We identified reputable sources of psychological consultation dialogues, including well-known mental health platforms and forums where professionals and users interact. These sources were chosen based on their high user engagement and the quality of their content.
    \item \textbf{Data Extraction}: Using web scraping techniques and API access where available, we extracted dialogues that covered a wide range of mental health issues, such as anxiety, depression, stress, relationship problems, and more. This extraction process ensured that we captured a variety of interaction types and user concerns.
    \item \textbf{Anonymization and Cleaning}: To maintain privacy and confidentiality, all personal identifiers were removed from the dialogues. The data was then cleaned to eliminate any irrelevant or sensitive information, ensuring that the dataset adhered to ethical standards for data usage.
    \item \textbf{Format Standardization}: The collected dialogues were formatted uniformly to facilitate ease of use in training and evaluation. Each dialogue was structured to include metadata such as the topic, user demographics (where available), and the context of the consultation.
\end{itemize}

The resulting dataset comprises thousands of dialogues, each providing a rich context for training and evaluating large language models in the domain of psychological consultation.

\subsection{Evaluation Metrics: GPT-4 as Judge}

Traditional evaluation metrics for natural language generation, such as BLEU and ROUGE, focus on lexical similarity and do not adequately capture the nuanced requirements of psychological consultation, such as empathy, relevance, and contextual appropriateness. To address this limitation, we employed GPT-4 as a judge for evaluating the performance of our model.

\subsubsection{Evaluation Framework}

The evaluation framework using GPT-4 involves several key components:

\begin{itemize}
    \item \textbf{Contextual Understanding}: GPT-4 assesses whether the model's responses accurately reflect an understanding of the user's context and concerns. This includes evaluating the relevance and appropriateness of the responses in the given context.
    \item \textbf{Empathy and Support}: The ability of the model to provide empathetic and supportive responses is crucial in psychological consultation. GPT-4 evaluates the tone and emotional intelligence of the responses, ensuring they offer genuine support and understanding.
    \item \textbf{Interactive Engagement}: Effective psychological consultation requires engaging the user in meaningful dialogue. GPT-4 judges the model's ability to ask relevant follow-up questions, provide insightful feedback, and maintain an interactive conversation flow.
    \item \textbf{Professionalism and Accuracy}: The accuracy of the information and advice provided is paramount. GPT-4 evaluates the correctness of the content and the professional quality of the responses, ensuring they align with accepted psychological practices and knowledge.
\end{itemize}

\subsubsection{Implementation of GPT-4 Evaluation}

To implement GPT-4 as an evaluator, we utilized a few-shot in-context learning approach. This involved the following steps:

\begin{itemize}
    \item \textbf{Prompt Design}: We designed specific prompts to guide GPT-4 in evaluating the dialogues based on the aforementioned criteria. These prompts included examples of high-quality responses and detailed instructions on what aspects to consider during evaluation.
    \item \textbf{Comparison and Scoring}: GPT-4 was tasked with comparing the model-generated responses against a set of predefined benchmarks and providing scores based on the evaluation criteria. Additionally, GPT-4 offered qualitative feedback to highlight areas of strength and improvement.
    \item \textbf{Iterative Refinement}: The evaluation process was iterative, with continuous refinement of prompts and evaluation criteria based on the feedback from GPT-4. This iterative approach ensured that the evaluation framework remained robust and aligned with the goals of psychological consultation.
\end{itemize}

By leveraging GPT-4's advanced natural language understanding capabilities, we were able to develop a comprehensive and nuanced evaluation framework that goes beyond traditional metrics, providing deeper insights into the performance and effectiveness of our prompt engineering techniques in enhancing LLM-based psychological consultation.

\begin{itemize}
    \item We collected a diverse and comprehensive dataset of psychological consultation dialogues from multiple reputable sources, ensuring a wide range of mental health issues and interaction types.
    \item We developed an innovative evaluation framework using GPT-4 as a judge, focusing on contextual understanding, empathy, interactive engagement, and professional accuracy.
    \item Our evaluation framework employs few-shot in-context learning with GPT-4, providing a robust and nuanced assessment of model performance, offering qualitative feedback, and guiding iterative refinement.
\end{itemize}

\section{Method}

Our method leverages advanced prompt engineering techniques to enhance the performance of large language models (LLMs) in the domain of psychological consultation. This section details the specific prompts we developed, the motivation behind these prompts, their input and output structures, and the rationale for their effectiveness.

\subsection{Prompt Engineering Motivation}

The motivation for our prompt engineering approach is rooted in the need to address several key challenges in AI-driven psychological consultation. Traditional LLMs often lack the nuanced understanding and empathetic response generation required for effective psychological support. Our goal is to create prompts that not only gather comprehensive user information but also encourage the LLM to generate emotionally intelligent and contextually appropriate responses. By designing layered, empathy-driven, and scenario-based prompts, we aim to enhance the LLM's ability to engage users in meaningful dialogues, provide accurate advice, and maintain a supportive tone throughout the interaction.

\subsection{Prompt Structure and Design}

Our prompt engineering approach involves creating a series of dynamic and layered prompts that guide the LLM through various stages of the consultation process. Each prompt is designed to fulfill a specific role, from initial information gathering to providing tailored advice and empathetic support.

\subsubsection{Initial Information Gathering Prompt}

The initial prompt is designed to open the conversation and gather broad information about the user's concerns. An example of this prompt is:

\begin{quote}
\textit{"Can you tell me more about what's been on your mind lately?"}
\end{quote}

\textbf{Input:} The input to this prompt is typically a user's brief description of their current mental state or specific concerns.

\textbf{Output:} The output is a detailed response from the user that provides context for further conversation.

\textbf{Significance:} This prompt is crucial as it sets the tone for the interaction and helps the LLM understand the user's primary concerns. It encourages users to share openly, which is essential for accurate and relevant advice generation.

\subsubsection{Context-Sensitive Follow-Up Prompts}

Following the initial information gathering, we employ context-sensitive prompts to delve deeper into the user's issues. Examples include:

\begin{quote}
\textit{"How have these thoughts affected your daily life?"}

\textit{"Have you noticed any patterns or triggers for these feelings?"}
\end{quote}

\textbf{Input:} The input here includes the user's previous responses, providing context for the LLM to generate more specific follow-up questions.

\textbf{Output:} The output is additional detailed information from the user, offering deeper insight into their mental health concerns.

\textbf{Significance:} These prompts help to uncover the root causes and specific triggers of the user's issues, facilitating a more targeted and effective consultation process.

\subsubsection{Empathy-Driven Prompts}

To ensure the LLM responds with appropriate empathy and support, we incorporate empathy-driven prompts such as:

\begin{quote}
\textit{"That sounds really challenging, can you share more about how you're coping?"}

\textit{"It's okay to feel this way, let's explore what might help you feel better."}
\end{quote}

\textbf{Input:} The input for these prompts is typically a user’s expression of distress or difficulty.

\textbf{Output:} The output is a supportive and empathetic response that validates the user's feelings and offers comfort.

\textbf{Significance:} These prompts are essential for maintaining an empathetic tone, which is crucial in psychological consultation. They help to build rapport and trust between the user and the AI, encouraging further sharing and engagement.

\subsubsection{Scenario-Based Prompts}

Scenario-based prompts are used to simulate real-life counseling situations, training the LLM to handle a variety of psychological issues. An example prompt might be:

\begin{quote}
\textit{"Imagine a user comes to you feeling overwhelmed by work stress. How would you guide them through this issue?"}
\end{quote}

\textbf{Input:} The input for this prompt includes detailed scenarios based on common psychological issues.

\textbf{Output:} The output is a step-by-step guidance or advice tailored to the specific scenario.

\textbf{Significance:} These prompts train the LLM to apply psychological principles in practical situations, improving its ability to offer relevant and actionable advice.

\subsection{Effectiveness of the Proposed Method}

The effectiveness of our prompt engineering method lies in its ability to create a structured yet flexible dialogue flow that mimics the nuances of human psychological consultation. By incorporating layered prompts that adapt based on user input, we ensure that the LLM gathers comprehensive information and provides responses that are both contextually relevant and emotionally supportive. The use of empathy-driven prompts enhances the LLM's ability to connect with users on an emotional level, while scenario-based prompts improve its practical application of psychological knowledge. This approach not only enhances the user experience but also ensures that the advice provided is accurate, empathetic, and tailored to the user's unique needs.

In summary, our method combines advanced prompt engineering techniques with the powerful capabilities of LLMs to deliver a more effective and empathetic psychological consultation service. The structured prompts guide the LLM through various stages of the consultation process, ensuring comprehensive information gathering, deep contextual understanding, and supportive engagement. This innovative approach holds significant promise for improving the accessibility and quality of mental health support provided by AI-driven systems.

\section{Experiments}

In this section, we present the experiments conducted to evaluate the effectiveness of our proposed prompt engineering method. We compared our approach with a baseline method and the Chain-of-Thought (CoT) prompting method on both ChatGPT and GPT-4 models. The experimental results demonstrate that our method significantly outperforms the alternatives in terms of response quality, empathy, and user engagement.

\subsection{Experimental Setup}

We conducted our experiments using the following models and methods:

\begin{itemize}
    \item \textbf{ChatGPT Baseline}: Standard prompts without specific tailoring for psychological consultation.
    \item \textbf{GPT-4 Baseline}: Standard prompts using the GPT-4 model.
    \item \textbf{Chain-of-Thought (CoT) Prompting}: A method where the LLM is guided to think step-by-step before generating a response.
    \item \textbf{Proposed Method}: Our layered and empathy-driven prompt engineering approach applied to both ChatGPT and GPT-4.
\end{itemize}

Each model was evaluated based on its performance in generating responses to a diverse set of psychological consultation scenarios. We measured the effectiveness of the responses using several criteria, including relevance, empathy, context understanding, and overall user satisfaction.

\subsection{Evaluation Metrics}

To quantitatively assess the performance of each method, we used the following metrics:

\begin{itemize}
    \item \textbf{Relevance}: The degree to which the response addresses the user's specific concerns.
    \item \textbf{Empathy}: The extent to which the response shows understanding and compassion for the user's feelings.
    \item \textbf{Context Understanding}: The ability of the model to maintain context and provide consistent advice throughout the conversation.
    \item \textbf{User Satisfaction}: Overall satisfaction of users based on feedback collected after each interaction.
\end{itemize}

\subsection{Results}

The results of our experiments are summarized in Table \ref{tab:results}. Our proposed method consistently outperformed the baseline and CoT methods across all metrics.

\begin{table}[h]
\centering
\caption{Comparison of Different Methods on Various Metrics}
\label{tab:results}
\begin{tabular}{|l|c|c|c|c|}
\hline
\textbf{Method} & \textbf{Relevance} & \textbf{Empathy} & \textbf{Context Understanding} & \textbf{User Satisfaction} \\ \hline
ChatGPT Baseline & 3.2 & 3.0 & 3.1 & 3.2 \\ \hline
GPT-4 Baseline & 3.5 & 3.4 & 3.6 & 3.5 \\ \hline
CoT Prompting & 3.8 & 3.7 & 3.9 & 3.8 \\ \hline
Proposed Method (ChatGPT) & 4.2 & 4.4 & 4.3 & 4.5 \\ \hline
Proposed Method (GPT-4) & 4.5 & 4.7 & 4.6 & 4.8 \\ \hline
\end{tabular}
\end{table}

\subsection{Analysis and Discussion}

The experimental results clearly indicate that our proposed method significantly enhances the performance of LLMs in psychological consultation tasks. The improvements in relevance, empathy, context understanding, and user satisfaction highlight the effectiveness of our layered, empathy-driven, and scenario-based prompts.

\subsubsection{Relevance}

Our method outperformed the baseline and CoT methods in terms of relevance. The dynamic prompts allowed the LLM to gather comprehensive information from users, resulting in more accurate and pertinent responses.

\subsubsection{Empathy}

The empathy-driven prompts significantly enhanced the model's ability to provide compassionate and supportive responses. This is crucial in psychological consultation, where users seek not only advice but also emotional support.

\subsubsection{Context Understanding}

The scenario-based prompts improved the LLM's ability to maintain context throughout the conversation. This consistency is essential for building trust and providing reliable advice in psychological consultations.

\subsubsection{User Satisfaction}

The overall user satisfaction was highest with our proposed method, indicating that users found the interactions more helpful and engaging. The combination of relevant, empathetic, and contextually accurate responses contributed to a better user experience.

\subsubsection{Verification of Effectiveness}

To further validate the effectiveness of our method, we conducted additional analysis on the feedback collected from users. The qualitative feedback corroborated our quantitative findings, with users frequently mentioning the improved empathy and relevance of the responses. This additional analysis confirms that our prompt engineering approach not only performs well on objective metrics but also meets the subjective needs of users seeking psychological support.

In conclusion, our experiments demonstrate that our proposed prompt engineering method significantly enhances the performance of LLMs in psychological consultation tasks. The layered, empathy-driven, and scenario-based prompts enable the models to provide more relevant, compassionate, and contextually accurate responses, leading to higher user satisfaction and better overall outcomes in mental health support.

\section{Conclusion}
In this study, we introduced an innovative prompt engineering approach to improve the performance of large language models (LLMs) in the context of psychological consultation. Our method involves layered prompts that adapt dynamically to user inputs, empathy-driven prompts that foster supportive interactions, and scenario-based prompts that simulate real-life counseling scenarios. Experimental comparisons with baseline methods and Chain-of-Thought (CoT) prompting on ChatGPT and GPT-4 models revealed that our approach excels in generating relevant, empathetic, and contextually accurate responses. The superior performance was corroborated by quantitative metrics and qualitative user feedback, underscoring the method's ability to enhance user engagement and satisfaction. These findings suggest that our prompt engineering techniques can significantly elevate the quality and accessibility of AI-driven psychological consultation, offering a scalable solution to meet the growing demand for mental health support.
\bibliographystyle{splncs04}
\bibliography{mybibliography}
\end{document}